\documentclass[letterpaper, 10 pt, conference]{ieeeconf}  

\IEEEoverridecommandlockouts                              
\usepackage[utf8]{inputenc}
\usepackage[T1]{fontenc}

\usepackage{graphicx} 
\usepackage{epsfig} 
\usepackage{amsmath} 
\usepackage{amssymb} 
\usepackage{subcaption}
\usepackage{glossaries}
\usepackage[detect-all]{siunitx}
\newacronym{fov}{FOV}{field of view}
\newacronym{pano}{pano}{panorama}
\newacronym{icp}{ICP}{Iterative closest point}
\newacronym{imu}{IMU}{Inertial Measurement Unit}
\newacronym{sota}{SOTA}{state-of-the-art}
\newacronym{lidar}{lidar}{Light Detection and Ranging}
\usepackage{xcolor}
\usepackage[breaklinks=true, colorlinks, bookmarks=true, citecolor=blue, urlcolor=blue, linkcolor=blue]{hyperref}

\newcommand{\acro}{\text{LLOL}}
\newcommand{\llol}{\acro}
\newcommand{\flio}{\text{Fast-LIO2}}
\newcommand{\RR}{\mathbb{R}}

\newcommand{\etal}{\textit{et al}. }

\title{\LARGE \bf
\llol: Low-Latency Odometry for Spinning Lidars
}

\author{Chao Qu, Shreyas S. Shivakumar, Wenxin Liu and Camillo J. Taylor
\thanks{We gratefully acknowledge the support of Distributed and  Collaborative Intelligent Systems and Technology Collaborative Research Alliance (DCIST) and C-BRIC, a Semiconductor Research Corporation Joint University Microelectronics Program, program cosponsored by DARPA.}
\thanks{C. Qu, S. S. Shivakumar, W. Liu and C. J. Taylor are with the GRASP Laboratory, School of Engineering and Applied Sciences,
University of Pennsylvania
{\tt\small \{quchao, sshreyas, wenxinl, cjtaylor\} @seas.upenn.edu}}%
}

\begin{document}

\maketitle
\thispagestyle{empty}
\pagestyle{empty}

\begin{abstract}
In this paper, we present a low-latency odometry system designed for spinning lidars.
Many existing lidar odometry methods wait for an entire sweep from the lidar before processing the data. 
This introduces a large delay between the first laser firing and its pose estimate.
To reduce this latency, we treat the spinning lidar as a streaming sensor
and process packets as they arrive. 
This effectively distributes expensive operations across time, 
resulting in a very fast and lightweight system with a much higher throughput and lower latency.
Our open source implementation is available at \url{https://github.com/versatran01/llol}.
\end{abstract}
\section{Introduction}

\gls{lidar} sensors are one of the primary sensing modalities in autonomous driving and robotics applications.
They work by sweeping one or more laser beams through the scene and calculate range by measuring time of flight.
This is apparent in mechanical \glspl{lidar} that continually spin a comb of lasers. 
However, solid-state \glspl{lidar} designed to minimize moving parts also involve redirecting a laser beam over time to cover a wider \gls{fov}.

Many existing lidar odometry systems wait for the \gls{lidar} to complete a full revolution and then process the resulting sweep to update ego-motion along with any associated scene representations.
Because of this, systems that combine \glspl{imu}, cameras and \glspl{lidar} often accept that \gls{lidar} is the slowest sensor in the system, 
and design algorithms based on that assumption~\cite{Zhang2015VisuallidarOA, Shan2021LVISAMTL, Zhao2021SuperOI}.
At first glance this is a valid assumption since a spinning \gls{lidar} usually rotates at \SI{10}{Hz}, which is a much lower rate compared to those of a typical IMU ($>$\SI{100}{Hz}) or camera ($>$\SI{20}{Hz}).
However, this data capture delay is an abstraction of the underlying mechanism of the \gls{lidar},
and is adopted only for ease of processing.
It ignores the streaming nature of the sensor and the fact that the raw \gls{lidar} packets usually arrive at a much higher rate ($>$\SI{600}{Hz}).
Recent works have also demonstrated that directly operating on lidar packets enables low-latency object detection, which is critical for autonomous driving~\cite{Frossard2020StrObeSO, Han2020StreamingOD}.

\begin{figure}[t]
\centering
\includegraphics[width=\linewidth]{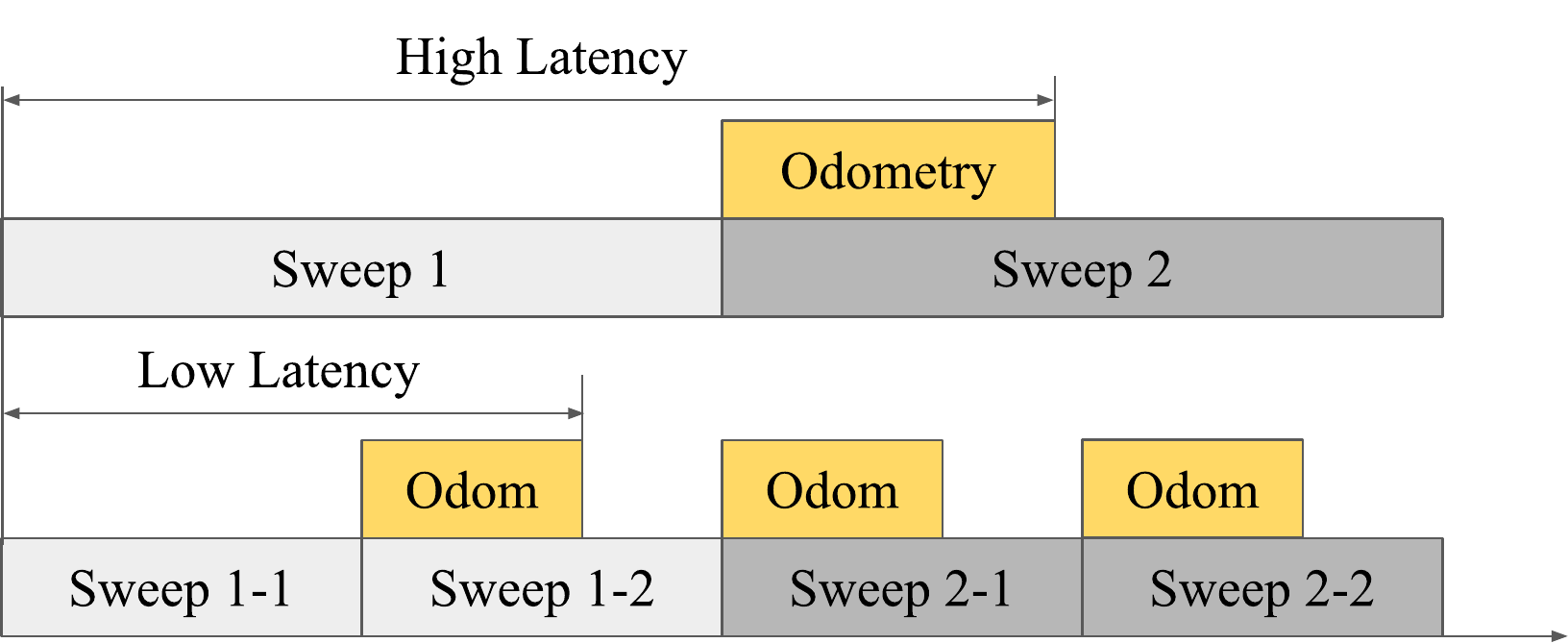}
\caption{Difference in latency when using full sweep (top) vs partial sweeps (bottom) in an odometry system. 
Gray blocks denote \gls{lidar} acquisition time,
and yellow blocks denote odometry runtime. 
We propose to process partial sweeps which allows the system
to produce odometry at a higher rate with lower latency.
}
\label{fig:latency}
\vspace{-5mm}
\end{figure}

In this paper we show how to exploit this continuous stream of data to significantly 
decrease latency and increase throughput for lidar odometry.
Fig.~\ref{fig:latency} shows the difference in latency of a standard lidar odometry 
and the one proposed in this paper. 
Consider the latency between the time of the laser firing at the beginning of the sweep
and the time that the sweep trajectory is estimated, we note that it has two components: 
the integration time of the \gls{lidar} and the runtime of the odometry. 
By this definition, a system that operates on a full sweep will have roughly twice the latency compared to one that operates on half a sweep.

Our proposed approach has two advantages.
Firstly it can process partial sweeps by employing a circular buffer of range measurements which drastically increases the rate at which we can produce new pose measurements, for example our approach can produce pose estimates at a rate of 80Hz from a lidar sensor with a sweep rate of 10Hz. 
Secondly our approach can process the resulting range information extremely efficiently on modern processor architectures without the need for GPU acceleration which minimizes the latency of those pose estimates. 
Both of these properties, increased throughput and decreased latency, 
are particularly important in contexts where the pose estimates are being used for closed loop control of dynamic platforms like quadrotors.

In the rest of this paper, we describe key elements of our system in Section~\ref{sec:method} and provide implementation details in Section~\ref{sec:detail}.
We then evaluate its accuracy and runtime by comparing to a state-of-the-art lidar odometry system~\cite{Xu2021FASTLIO2FD} in Section~\ref{sec:result}. 
We establish that by using partial sweeps we can increase the throughput substantially while lowering the effective latency. 
We show that by adopting a data-oriented design with array-based data structures, 
our system exhibits almost linear scalability in terms of multi-threading performance.
Unfortunately, many popular datasets only provide \gls{lidar} data pre-packaged as point clouds from full sweeps. 
Therefore, we also collected our own dataset to further validate our method. 
Our system, named LLOL, is open sourced and available to the community.

\section{Related Work}



In this section, we review related work on lidar-(inertial) odometry with a particular focus on systems that have been
demonstrated to run in real time on embedded platforms.

\subsection{Lidar Odometry}

In a seminal paper Zhang \etal proposed LOAM, a feature based lidar odometry designed to work on embedded platforms, 
with loose IMU integration and a discrete trajectory representation ~\cite{Zhang2014LOAMLO}. 
Shan \etal improved LOAM for ground robots by leveraging an initial segmentation procedure to detect ground planes~\cite{Shan2018LeGOLOAMLA}. 
LOAM was further extended by Lin \etal to work with limited \gls{fov} LiDARs,
supporting irregular sampling of feature points and enhanced outlier rejection.
Liu \etal proposed a method of local bundle adjustment and a temporal sliding window that allows for real-time performance while also improving odometry, and use LOAM as a front end in their validation~\cite{Liu2021BALMBA}. 
Elastic-Fusion, SuMa and MARS use a surfel based map to improve performance, and allow representation of larger scale environments~\cite{Park2018ElasticLF, Park2020ElasticityMC, Behley2018EfficientSS, Quenzel2021RealtimeM6}. 
Palieri \etal proposed LOCUS, a multi-sensor lidar-centric system for odometry and mapping that is robust to intermittent sensor degradation, and deployed it on both legged and wheeled robot platforms. 
Yokozuka \etal proposed LITAMIN~\cite{Yokozuka2020LiTAMINLT}, which is a light-weight lidar odometry. 
They adopt a voxelization step to reduce the number of points used in ICP and employ a modified GICP~\cite{Segal2009GeneralizedICP} cost that avoids expensive eigen-decomposition computations commonly used with the point-to-plane metric. 
LITAMIN2 further improves real-time performance by matching two distributions via a symmetric KL-divergence cost~\cite{Yokozuka2021LiTAMIN2UL}. 



\subsection{Lidar-inertial Odometry}

Neuhaus \etal propose MC2SLAM, which is a tightly-coupled odometry system that is designed with real-time performance in mind. 
They report a runtime of approximately \SI{60}{ms}, 
excluding an intermittent pose graph optimization step that requires an additional \SI{108}{ms} every five frames. 
In~\cite{Li2021TowardsHS}, the authors propose LiLi-OM to address the irregular scan pattern of the Livox \gls{lidar} similar to~\cite{Lin2020LoamLA}. 
They perform tightly coupled lidar-inertial fusion with a keyframe-based sliding window optimization. 
They report a runtime performance of roughly \SI{85}{ms} per sweep on Velodyne lidar data. 
LIO-SAM~\cite{Shan2020LIOSAMTL} improves upon~\cite{Ye2019TightlyC3} and uses a window of local sweeps instead of matching individual lidar sweeps to a global map. 
They report an improvement in efficiency and accuracy, and a runtime performance of roughly \SI{50}{ms}-\SI{100}{ms} per sweep on a laptop CPU.
Xu \etal proposed Fast-LIO and \flio, which we use in our quantitative comparisons given their focus on real-time performance and support for similar hardware as our method ~\cite{Xu2021FASTLIOAF, Xu2021FASTLIO2FD}. 
Fast-LIO2 achieves real- performance on embedded CPUs via the use of an incremental KD Tree and tightly-couples \gls{imu} data.
Perhaps the most similar work to ours is UPSLAM~\cite{Cowley2021UPSLAMUO}, 
which uses a union of depth panoramas~\cite{Taylor2015MappingWD} as its map representation.
It is also a full-fledged SLAM system that includes loop closure.
However, UPSLAM requires a GPU since it utilizes all points from a lidar sweep.




\section{Method}\label{sec:method}

A spinning \gls{lidar} sensor produces data in packets which can be decoded by a driver to produce a final point cloud.
It can also be converted into a spherical coordinate system resulting in a range image.
In this work, we define a lidar \textbf{sweep} to be either a point cloud or a range image generated from a complete revolution of the sensor.

Our system consumes lidar packets directly and 
accumulates a partial sweep as a range image internally~(Fig.~\ref{fig:input-sweep}).
Compared with structured point clouds, range images are more compact and memory efficient.
Once the partial sweep reaches a desired integration time, 
the odometry system processes and produces a pose estimate.
For ease of illustration, we assume a full sweep as input in this section. 
We address how to adapt our system to partial sweeps in section~\ref{sec:detail}.

\subsection{Feature Detection}\label{sec:method/grid}

\begin{figure}
\vspace{6 px}
\centering
\begin{subfigure}[b]{\linewidth}
        \centering
        \includegraphics[width=\linewidth]{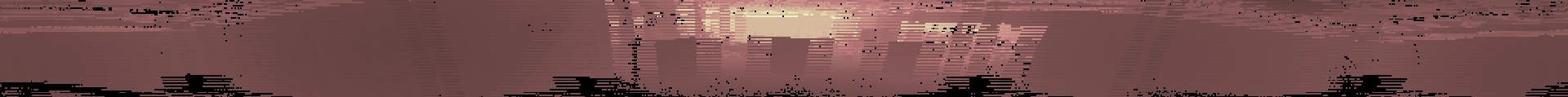}
        \caption{Input sweep}
        \label{fig:input-sweep}
    \end{subfigure}
    \hfill
    \begin{subfigure}[b]{0.49\linewidth}
        \centering
        \includegraphics[width=\linewidth]{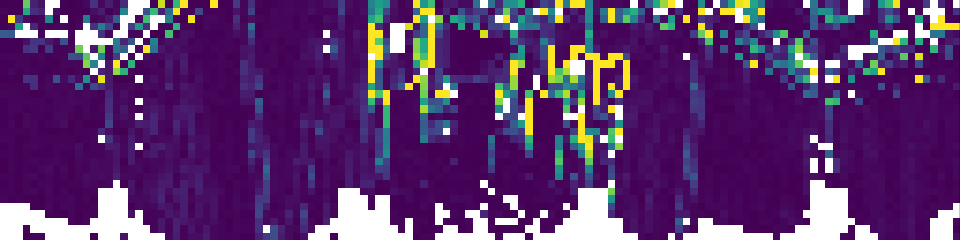}
        \caption{Cell smoothness}
        \label{fig:grid-smooth}
    \end{subfigure}
    \hfill
    \begin{subfigure}[b]{0.49\linewidth}
        \centering
        \includegraphics[width=\linewidth]{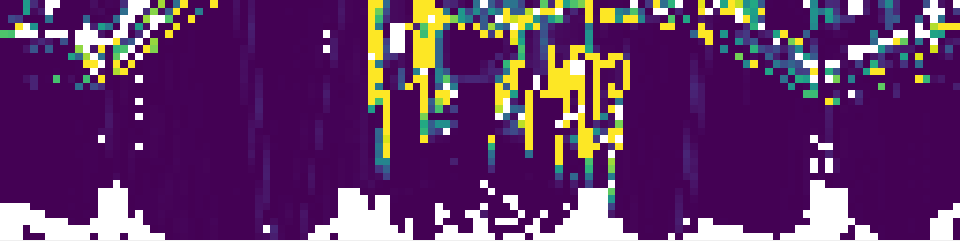}
        \caption{Cell variance}
        \label{fig:grid-var}
    \end{subfigure}
    \hfill
    \begin{subfigure}[b]{0.49\linewidth}
        \centering
        \includegraphics[width=\linewidth]{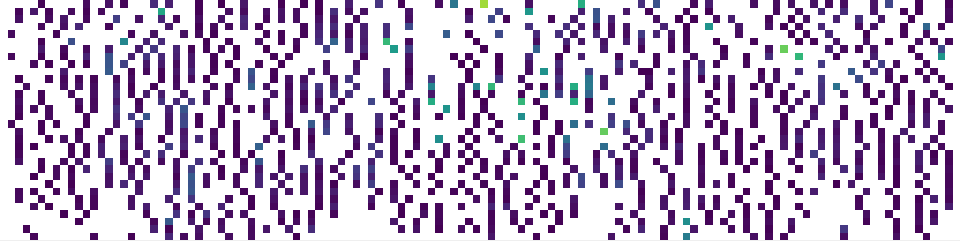}
        \caption{Match candidates}
        \label{fig:grid-filter}
    \end{subfigure}
    \hfill
    \begin{subfigure}[b]{0.49\linewidth}
        \centering
        \includegraphics[width=\linewidth]{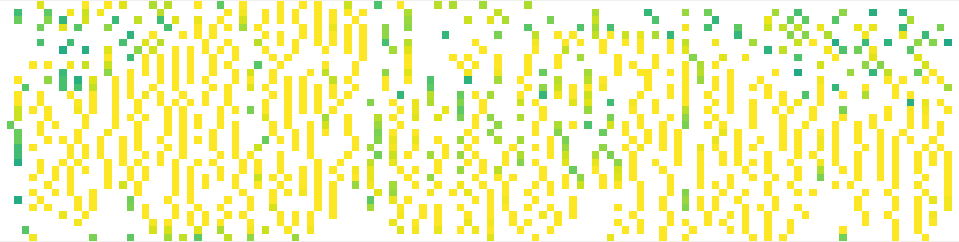}
        \caption{Matched cells}
        \label{fig:grid-match}
    \end{subfigure}
\caption{Various stages in the feature detection process. 
Brighter color denotes higher value in both colormaps.}
\label{fig:grid}
\vspace{-5px}
\end{figure}
\begin{figure}
\centering
\begin{subfigure}[b]{\linewidth}
    \centering
    \includegraphics[width=\linewidth]{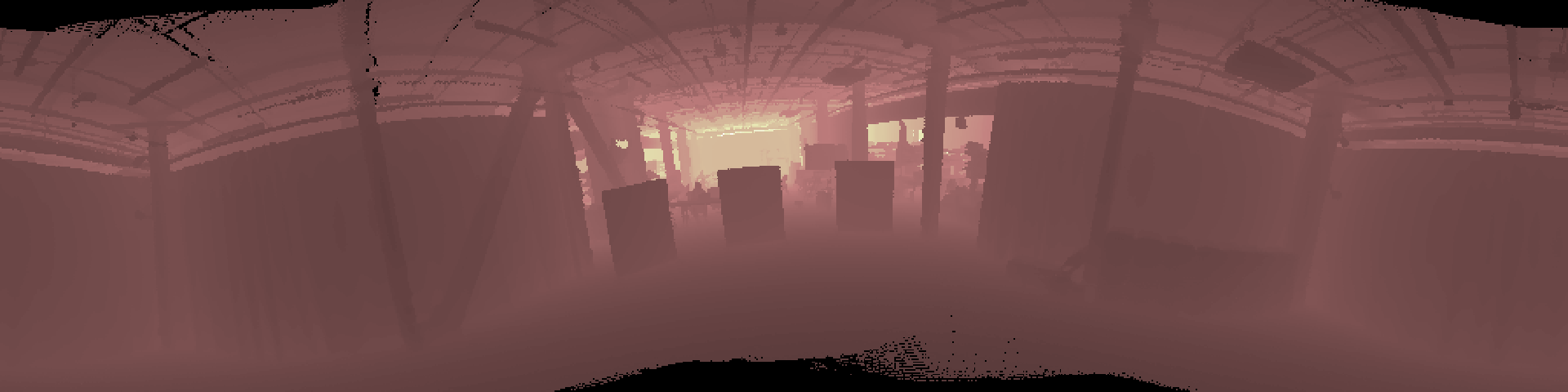}
\end{subfigure}
\hfill
\begin{subfigure}[b]{\linewidth}
    \centering
    \includegraphics[width=\linewidth]{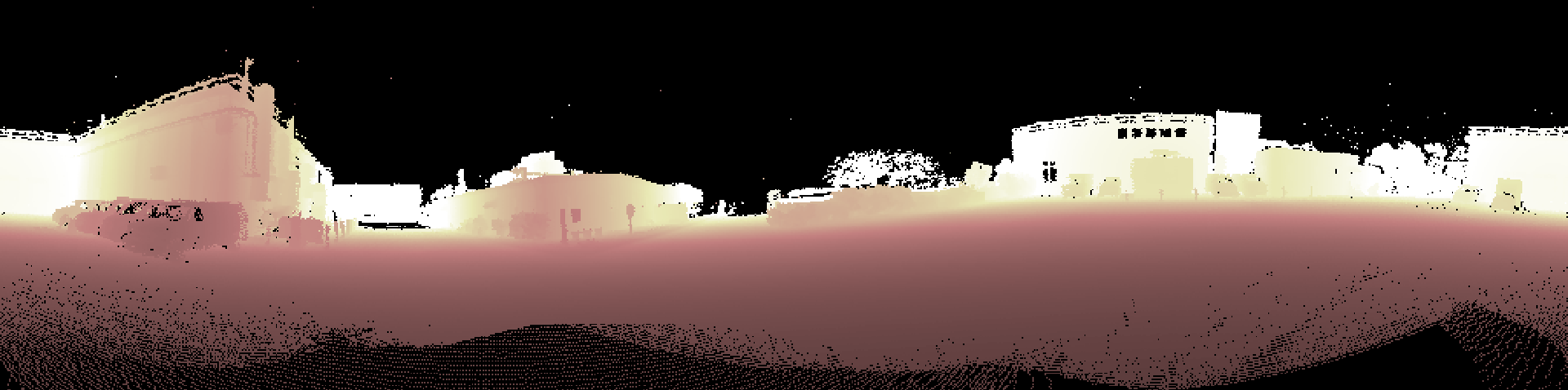}
\end{subfigure}
\caption{Depth panoramas generated by our system for both indoor (top) and outdoor (bottom) scenes using the same resolution ($256\times 1024$) with a 90\textdegree~vertical \gls{fov}.
}
\label{fig:pano}
\vspace{-5mm}
\end{figure}

The input to our system is a range image from which we can recover the 3D position of each pixel.
There is usually too much information to process, as a typical lidar sweep consists of $64\times 1024 = 65536$ data points. 
To reduce the amount of information, we divide the sweep into a grid of 2D cells and compute smoothness~(\ref{fig:grid-smooth}) and variance~(\ref{fig:grid-var}) for each cell.
Similar to~\cite{Shan2018LeGOLOAMLA}, smoothness $s$ is defined as
\begin{align}
s = \frac{1}{|C|}\left\vert \sum_{n\in C} \frac{r_n}{r_m} - 1 \right\vert
\end{align}
where $r_n$ is the range of pixel $n$, $r_m$ is the range of the midpoint and $|C|$ is the number of valid pixels in cell $C$.

We then filter the entire grid by a maximum smoothness and variance threshold to identify planar regions in the sweep. 
An optional non-minimum suppression can be applied to further reduce the number of cells
while maintaining a uniform distribution across the grid. 
We represent each remaining cell~(\ref{fig:grid-filter}) with a multivariate Gaussian  with mean $\mu^{\mathcal{S}} \in \RR^3$ and covariance $\Sigma^{\mathcal{S}} \in \RR^{3 \times 3}$ which captures the statistics of the 3D positions of the points in that cell. 
These quantities will be used for data association and optimization.
This feature detection stage is similar to the voxelization step in LITAMIN2~\cite{Yokozuka2021LiTAMIN2UL} but is carried out in the 2D image space.

\subsection{Local Map Representation}\label{sec:method/pano}

Following UPSLAM~\cite{Cowley2021UPSLAMUO}, we employ a depth \gls{pano}~\cite{Taylor2015MappingWD} as our local map.
A \gls{pano} is similar to a sweep, but with higher resolution and represents the 3D structure at one point along the sensors trajectory.
It is built by fusing multiple localized and motion-corrected sweeps.
Compared with 3D map representations like point clouds or voxel/surfel grids, 
a depth \gls{pano} has a fixed storage complexity.
The advantages include low memory footprint, 
linear access patterns and adaptive spatial resolution.
As shown in Fig.~\ref{fig:pano}, 
the same \gls{pano} size can be used to model both indoor and outdoor scenes. 
Whereas with 3D representations, one has to choose a suitable spatial resolution to voxelize the workspace,
which results in high variance in runtime efficiency.

Due to its fixed viewpoint and spherical projection model, 
our \gls{pano} map is susceptible to occlusions and limited horizontal \gls{fov}.
In practice, we found this to be less of a problem since most of our robots move in open space with planar motions.
Note that a graph of \gls{pano}s can be used for pose graph optimization to reduce drift~\cite{Cowley2021UPSLAMUO}, or fed into other loop closure systems~\cite{Kim2018ScanCE, Wang2020IntensitySC}.

\subsection{Local Trajectory Representation}\label{sec:method/traj}

At any given time, we maintain a local trajectory of the current sweep, 
which has a time span of $[t_0, t_1)$.
It consists of a set of discrete states, each for a single column of cells in the grid.
Given the laser firing time $\Delta t$, we can recover the column time 
by $t_c = t_0 + c \cdot \Delta t$.
Each state $X^\mathcal{P}_{\mathcal{S}_t}$ consists of the position, velocity and orientation of the sensor at time $t \in [t_0, t_1)$ with respect to the current \gls{pano} frame
\begin{align}
X^\mathcal{P}_{\mathcal{S}_t} = \begin{bmatrix} p^\mathcal{P}_{\mathcal{S}_t} & v^\mathcal{P}_{\mathcal{S}_t} & R^\mathcal{P}_{\mathcal{S}_t} \end{bmatrix}
\end{align}
Given a new sweep, we start from the last state of the previous sweep and propagate forward in time using the available \gls{imu} readings to obtain an initial guess. 
This trajectory estimate is then used to resolve the locations of the features in the sweep.


\subsection{Projective ICP} \label{sec:icp}

\begin{figure}
\vspace{6px}
\centering
\includegraphics[width=\linewidth]{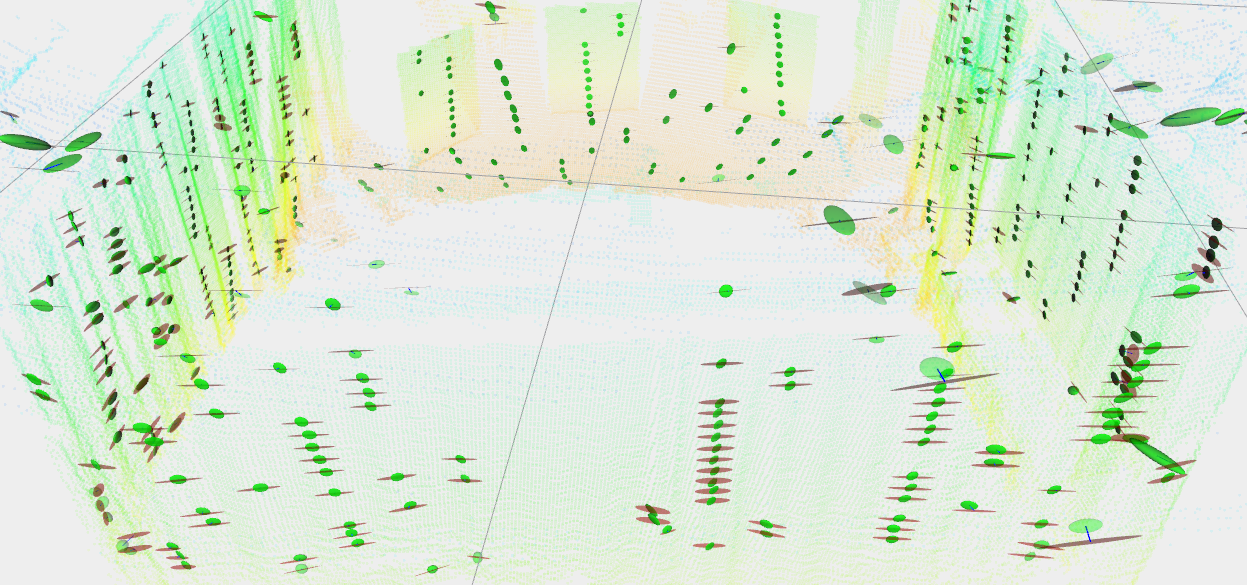}
\caption{Data association between sweep and pano.
Red and green ellipsoids depict the normal distributions of grid cells 
and their matching windows in the \gls{pano}.}
\label{fig:icp-match}
\vspace{-5mm}
\end{figure}

Given a set of match candidates (planar cells from~\ref{sec:method/grid}), 
we can estimate the pose of the current sweep by aligning it to the \gls{pano} via an \gls{icp} procedure.

\subsubsection{Data Association}

We find correspondences by projective data association~\cite{Serafin2015NICPDN}, 
where points in the sweep frame $\mathcal{S}$ are transformed into the \gls{pano} frame $\mathcal{P}$
\begin{align}\label{eqn:proj}
u^\mathcal{P} = \pi(T^{\mathcal{P}}_{\mathcal{S}} \mu^{\mathcal{S}})
=\pi(R^{\mathcal{P}}_{\mathcal{S}} \mu^{\mathcal{S}} + p^\mathcal{P}_\mathcal{S}),
\end{align}
Here $\mu^\mathcal{S}$ is the mean position of the cell from the sweep
and $u^\mathcal{P}$ is its target pixel coordinate in the \gls{pano}.
$\pi(\cdot)$ is the spherical projection function, and
$T^{\mathcal{P}}_{\mathcal{S}} \in \mathbb{SE}(3)$ 
is the rigid transformation from the sweep to the pano frame.
We only accept matches that pass the following checks: 
1) the target pixel must lie within the \gls{pano},
2) the depths of the target pixel and sweep point must be close enough, and
3) more than half of the pixels within a window
around the target must be valid.
These criteria will eliminate most of the outliers due to occlusion.
The remaining ones can be handled by either a robust loss function or $\chi$-squared test.
We summarize the target pixel with a multivariate Gaussian with mean $\mu^\mathcal{P} \in \RR^3$ and covariance $\Sigma^\mathcal{P} \in \RR^{3 \times 3}$ using all valid pixels in the window. 

\subsubsection{Pose Optimization}

Once the correspondences are established, as shown in Fig.~\ref{fig:icp-match}, 
we proceed to the registration step.
We employ the GICP~\cite{Segal2009GeneralizedICP} cost function in the optimization procedure,
\begin{align}
r^2 = \| \mu^\mathcal{P} - T^\mathcal{P}_\mathcal{S} \mu^\mathcal{S} \|^2_{\Sigma}, \quad 
\Sigma = \Sigma^\mathcal{P} + R^\mathcal{P}_\mathcal{S} \Sigma^\mathcal{S} R^\mathcal{S}_\mathcal{P}
\end{align}
where $\| \cdot \|_\Sigma$ is the Mahalanobis distance and
$\Sigma^\mathcal{P}$ and $\Sigma^\mathcal{S}$ are the covariance matrices in the sweep and pano frame respectively. 
When \gls{imu} data is available, 
we add an \gls{imu} preintegration factor~\cite{Forster2017OnManifoldPF} to constrain the start and the end state of the local trajectory.
The error state parameters we wish to optimize represent a small correction $\Delta T^\mathcal{P'}_\mathcal{P}$ 
in the \gls{pano} frame such that the new state $X^\mathcal{P'}_\mathcal{S}$ can be updated via 
\begin{align}\label{eq:update}
T^\mathcal{P'}_\mathcal{S} = \Delta T^\mathcal{P'}_\mathcal{P} \cdot T^\mathcal{P}_\mathcal{S}, 
\quad 
v^\mathcal{P'}_\mathcal{S} = v^\mathcal{P}_\mathcal{S} + \Delta p^\mathcal{P'}_\mathcal{P} / \Delta t
\end{align}
where $\Delta p^\mathcal{P'}_\mathcal{P}$ is the translation part of $\Delta T^\mathcal{P'}_\mathcal{P}$ and $\Delta t = t1 - t0$.
After a few iterations of the Levenberg-Marquardt procedure, 
the resulting update is applied to the starting state of the sweep trajectory.
We then repropagate the entire trajectory and update correspondences for the next round of \gls{icp}, similar to VICP~\cite{Hong2010VICPVU}.
\subsection{Map Update}

\subsubsection{Depth Fusion}
\begin{figure}[t]
\vspace{6px}
\centering
\includegraphics[width=\linewidth]{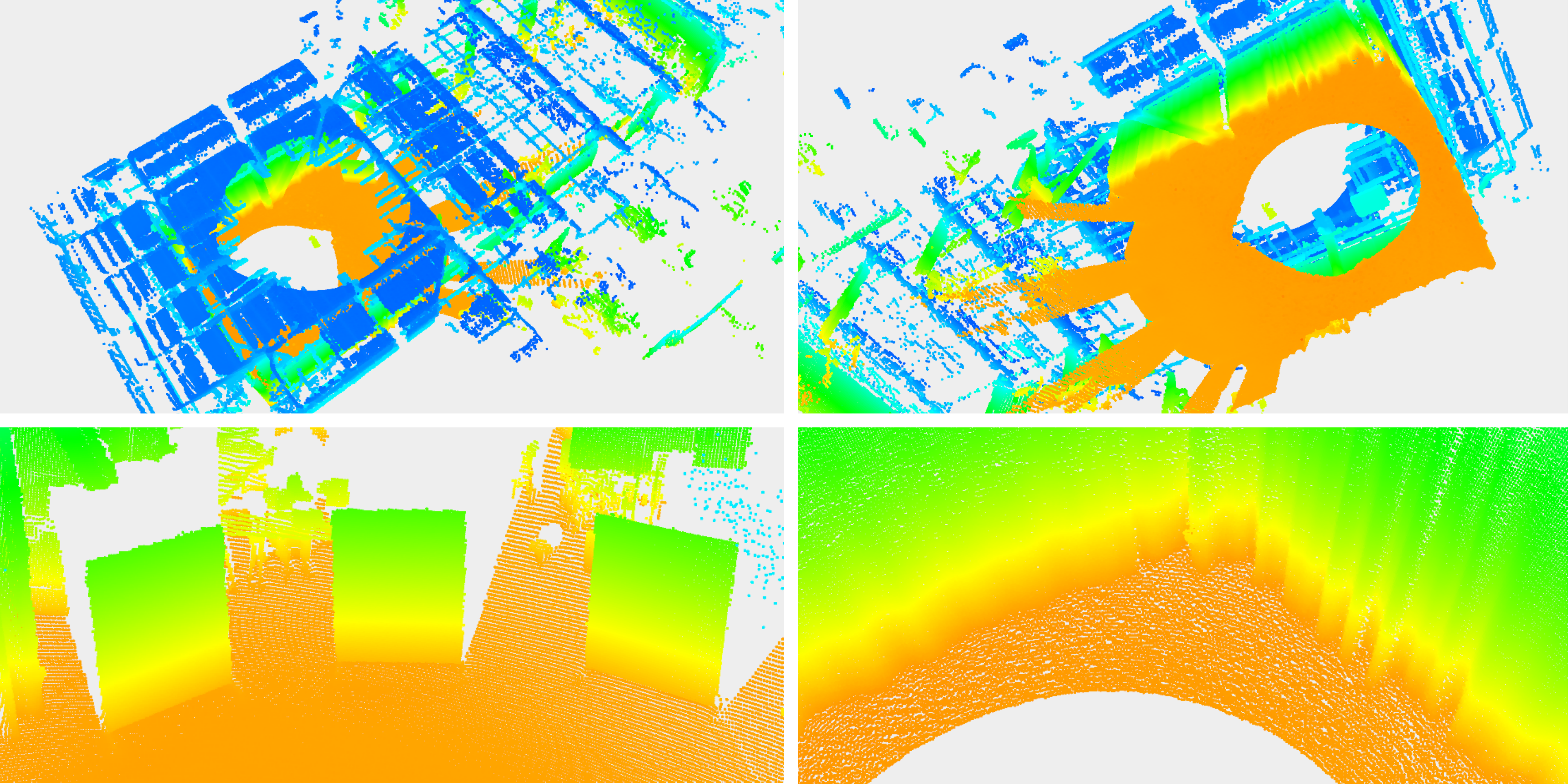}
\caption{An indoor space (top) mapped with our lidar odometry. 
The points are generated by back-projecting a depth panorama.
Details of the scene (bottom) are faithfully preserved by the depth panorama.}
\label{fig:cloud-indoor}
\vspace{-5mm}
\end{figure}

Once the \gls{icp} procedure has converged, we can undistort the current sweep using the optimized local trajectory and update the \gls{pano}.
This is achieved by projecting every pixel from the sweep to the \gls{pano} using Eq.~\ref{eqn:proj}.
Although in principle one could maintain a probabilistic depth filter for each pixel and perform recursive Bayesian updates~\cite{Forster2014SVOFS} given the new depth, 
we found that in practice, the similarity filter used in~\cite{Cowley2021UPSLAMUO} worked exceptionally well with very low computation and storage overhead.
We keep a counter $k$ alongside the depth value for each pixel in the \gls{pano}.
If the new depth $d'$ is close enough to the stored depth $d$, 
we do a weighted average update and increment the counter by 1, 
saturating at $k_\mathrm{max}$. 
\begin{align}
d \leftarrow \frac{k \cdot d + d'}{k + 1}, \quad 
k \leftarrow \min(k_{\mathrm{max}}, k + 1)
\end{align}
We otherwise treat the new depth as an outlier and simply decrements the counter by 1 without modifying the current value.
If $k$ reaches 0, we replace $d$ with $d'$.
This simple depth fusion strategy produces smooth depth surfaces as is evident in Fig.~\ref{fig:cloud-indoor}.
It also makes our map resilient to ephemeral objects, 
while still being able to adapt to long-term environment changes. 
For example, a car driving across the lidar will typically leave a trail of ghost points in a point cloud map,
but it will not be registered into the \gls{pano} unless it has stopped moving for a long enough time.

\subsubsection{Rendering New Panoramas}

As the sensor moves away from the \gls{pano}, 
the number of matches between the two will gradually decrease.
This is due to the projective nature of range images, 
where data association is affected by occlusions and non-overlapping \glspl{fov}.
Therefore, we need to move the \gls{pano} to a new position in order to 
retain a healthy number of matches for \gls{icp}.

The decision on whether to relocate the \gls{pano} is made by looking at the matching quality $Q$, which is defined as the ratio between the number of final matches and matching candidates.
This number essentially reflects the amount of non-occluding overlap between the sweep and the \gls{pano}.
Once $Q$ drops below a certain threshold (0.9 by default), we will render a new \gls{pano} at the current location.
If \gls{imu} measurements are available and motion in roll and pitch is limited, the new \gls{pano} will be placed in a gravity aligned frame.
Since we do not have dense depth information in the new location, 
we do a forward warping from the old \gls{pano} to the new one.
This will naturally result in many holes in the new \gls{pano}, 
but this can be alleviated by immediately incorporating the current sweep.
The rendering process is executed at a much lower rate, 
since it is only triggered when the two viewpoints start to deviate from each other.

\section{Implementation}\label{sec:detail}

\begin{figure}[t]
\vspace{6px}
\centering
\includegraphics[width=\linewidth]{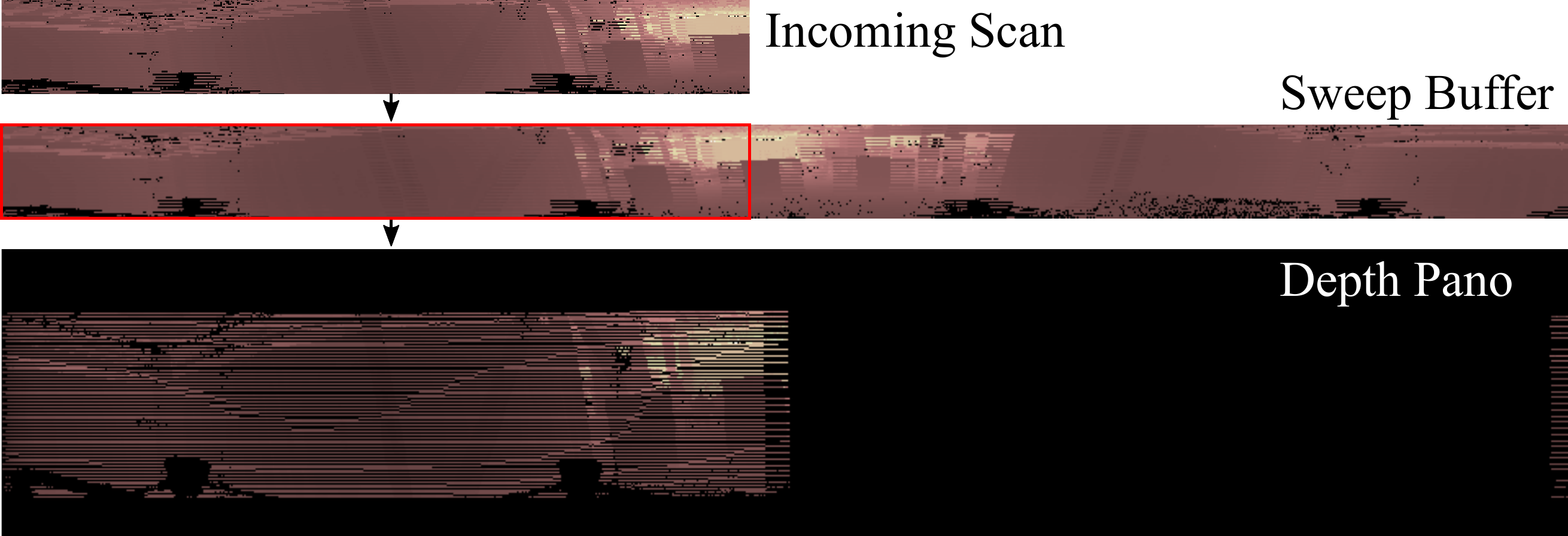}
\caption{Illustration of how partial sweeps are handled.
The incoming partial sweep is pushed into a sweep buffer.
The part being ejected from the buffer is added to the pano.}
\label{fig:partial}
\vspace{-5mm}
\end{figure}

\subsection{Processing Partial Sweeps}\label{sec:detail/partial}

To handle partial sweeps, we maintain a circular buffer for a full sweep, 
and push incoming partial sweeps into the buffer as they arrive.
The part of the buffer being ejected is 
immediately fused with the pano using the current state estimate.
This is shown in Fig.~\ref{fig:partial}, 
where the ejected part is outlined in red.
However, performing \gls{icp} with a partial sweep is challenging because the smaller the size, 
the more likely it is to encounter degeneracy.
Therefore, we use the entire sweep in the buffer for this task, which has 360\textdegree~\gls{fov}.
This adaptation is key to minimizing latency.
For example, if we process half of a sweep at a time, 
this reduces the sensor integration time to \SI{50}{ms} (assuming a \SI{10}{Hz} spinning rate).
It also halves the feature detection and pano update time, 
as they only need to process half of the data.
The \gls{icp} time remains unchanged, 
since it always tries to align the full sweep in the buffer to the \gls{pano}.
This essentially increases the odometry frequency to \SI{20}{Hz}.
One could further divide the sweep into smaller chunks, 
but there exists a lower bound on how narrow a partial sweep can be handled without dropping data,
which is subject to the ICP processing time.
In practice, our system can handle 1/8th of a sweep which produces odometry at \SI{80}{Hz}.

\subsection{Memory Layout and Parallelization}

Our implementation follows a data-oriented design~\cite{Richard2018DOD} 
and is optimized for cache-locality and parallelism.
We only use array-based data structures and pre-allocate all storage ahead of time.
There is no dynamic allocation occurring at runtime within our system, 
which makes it favorable for embedded applications.
Specifically, we only need to allocate enough storage for a full sweep, a feature grid, 
two \glspl{pano} and various arrays for the solver and the local trajectory.
Together they require no more than 5-10MB of storage.

The projective data association is also significantly faster
compared to systems that leverage even a carefully designed KD-tree~\cite{Cai2021ikdTreeAI}.
A tree data structure requires frequent node allocation and pointer chasing 
during search and insertion.
Without advanced memory management or allocator support, 
almost every node access is a cache miss, 
which severely hinders performance on modern processors.
On the other hand, the projection operation is constant time, 
which is conceptually similar to a hash-table lookup.

Finally, almost every step of our system can be parallelized, 
since we mostly operate on images.
We choose a single row in both the sweep and the pano as our atomic work unit, 
and use a task-based threading library~\cite{Voss2019ProTBB} for scheduling.
A finer grain size of a single pixel is possible, 
but it requires additional synchronization overhead to avoid false-sharing.
Rendering a new \gls{pano} is carried out in the background via double buffering 
($<$\SI{10}{ms} in a single thread on Intel CPU), 
which does not interfere with the odometry.
\section{Results}\label{sec:result}

We evaluate both the accuracy and runtime performance of our proposed system.
For comparison we choose \flio~\cite{Xu2021FASTLIO2FD}, 
which is a state-of-the-art \gls{lidar}-inertial odometry with an open-source implementation\footnote{https://github.com/hku-mars/FAST\_LIO}.
We use their default configuration for Ouster \gls{lidar}\footnote{https://ouster.com/products/os1-lidar-sensor} 
with slightly tuned imu noise parameters.
For our system, unless otherwise specified, 
we use a pano of size $256\times 1024$, 
and a cell size of $2\times 16$.

\subsection{Odometry Accuracy}

\begin{figure}[t]
\vspace{6px}
\begin{subfigure}[b]{0.49\linewidth}
    \centering
    \includegraphics[width=\linewidth]{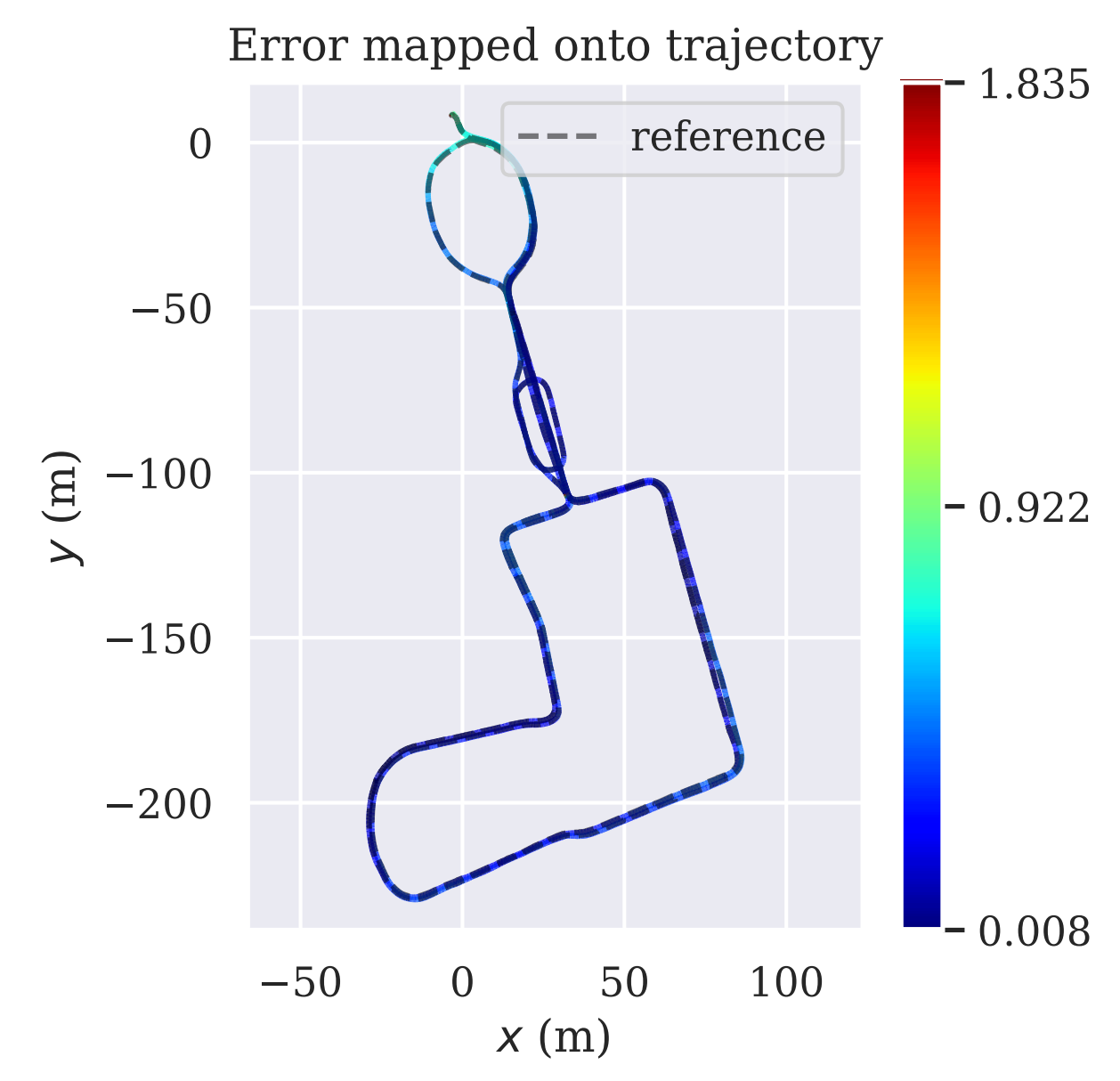}
    \caption{\flio}
    \label{fig:nc01-flio}
\end{subfigure}
\begin{subfigure}[b]{0.49\linewidth}
    \centering
    \includegraphics[width=\linewidth]{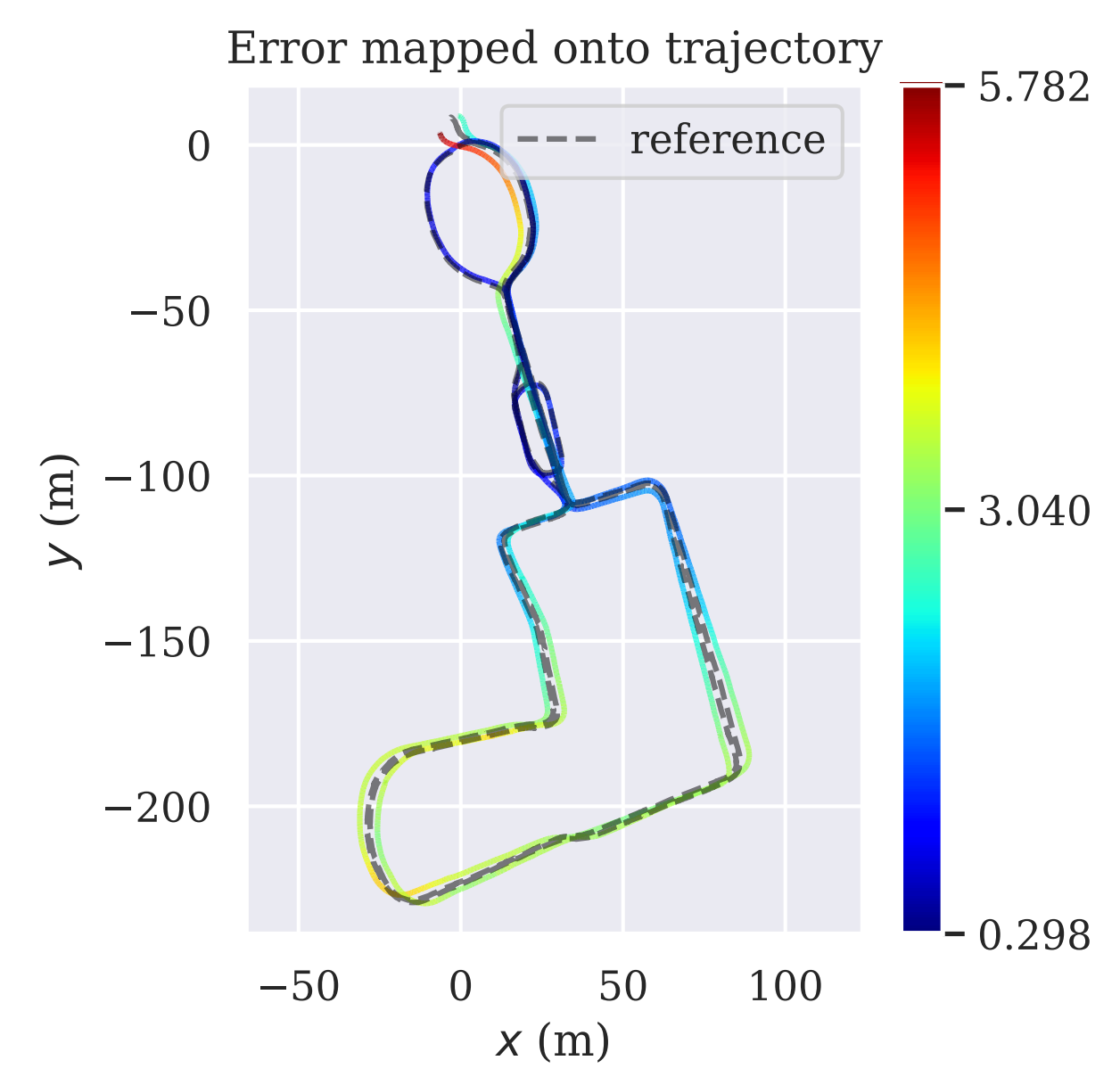}
    \caption{\llol}
    \label{fig:nc01-llol}
\end{subfigure}
\caption{Trajectory error plots of \flio~and \llol~on Newer College dataset sequence 1.}
\label{fig:nc01}
\vspace{-5mm}
\end{figure}
\begin{table}[t]
\vspace{6px}
\centering
\caption{Translational APE and RPE on NC01.}
\label{tab:nc01}
\begin{tabular}{ccccccc}
\hline
\multicolumn{3}{c}{Newer College Sequence 1} & \multicolumn{2}{c}{\flio} & \multicolumn{2}{c}{\llol} \\
Segment & $t_\mathrm{start}$ & $t_\mathrm{end}$ & APE & RPE & APE & RPE \\
\hline \hline 
1    &  25    & 340  & 0.363      & 0.214     & 0.396    & 0.283   \\
2    &  340   & 705  & 0.165      & 0.112     & 0.499    & 0.214   \\
3    &  705   & 1115 & 0.277      & 0.171     & 0.706    & 0.250   \\
4    &  1115  & 1500 & 0.311      & 0.121     & 0.912    & 0.238   \\
\hline
\end{tabular}
\vspace{-5mm}
\end{table}

\begin{table}[t]
\centering
\caption{Translational APE and RPE on NC05.}
\label{tab:nc05}
\begin{tabular}{ccccccc}
\hline
\multicolumn{3}{c}{Newer College Sequence 5} & \multicolumn{2}{c}{\flio} & \multicolumn{2}{c}{\llol} \\ 
Segment & $t_\mathrm{start}$ & $t_\mathrm{end}$ & APE & RPE & APE & RPE \\ 
\hline \hline 
1       & 25  & 100 & 0.104     & 0.172    & 0.130    & 0.328    \\
2       & 100 & 200 & 0.136     & 0.206    & 0.183    & 0.319   \\
3       & 200 & 300 & 0.102     & 0.122    & 0.178    & 0.258   \\ 
4       & 300 & 400 & 0.162     & 0.119    & 0.262    & 0.288   \\
\hline 
\end{tabular}
\end{table}

\subsubsection{Newer College Dataset}

As our system consumes lidar packets, this limits our choice of publicly available datasets to evaluate on.
The Newer College dataset~\cite{Ramezani2020TheNC} provides raw lidar packets along with ground truth trajectories.
We evaluate on two sequences: \textit{01-short-experiment (NC01)} and \textit{05-quad-with-dynamics (NC05)},
both collected with a handheld sensor platform at normal walking pace.
Note that \textit{NC05} features 4 loops with increasingly aggressive motion.
Within each sequence, the sensor is stopped at various locations for a short while.
This creates well-defined segments which allows for sub-trajectory evaluation.
We report results on these segments in Table~\ref{tab:nc01} and~\ref{tab:nc05},
and visualize the full trajectory of \textit{NC01} in Fig.~\ref{fig:nc01}.

We report the translational part of Absolute Pose Error (APE) and Relative Pose Error (RPE) metrics, 
both generated using the \textit{evo}~\cite{grupp2017evo} package. 
APE measures the difference in pose between the reference and the estimated trajectory. 
We use Umeyama alignment~\cite{umeyama1991least} as a pre-processing step to align the trajectories for all methods. 
RPE compares pairwise relative pose within a small window (\SI{1}{m}), 
which can be used to quantify odometry drift. 

\flio~outperforms \llol~in both sequences. 
This is because part of the Newer College dataset was collected in a park with many trees and few man-made structures. 
This results in reduced planar features for our system.
In addition, the sensor went under tree canopies several times, the heavy occlusion triggered more frequent rendering of the \gls{pano}, 
thus causing extra drift.

\subsubsection{Our Own Dataset}

\begin{figure*}
\vspace{6px}
\centering
    \begin{subfigure}[b]{0.245\linewidth}
        \centering
        \includegraphics[width=\linewidth]{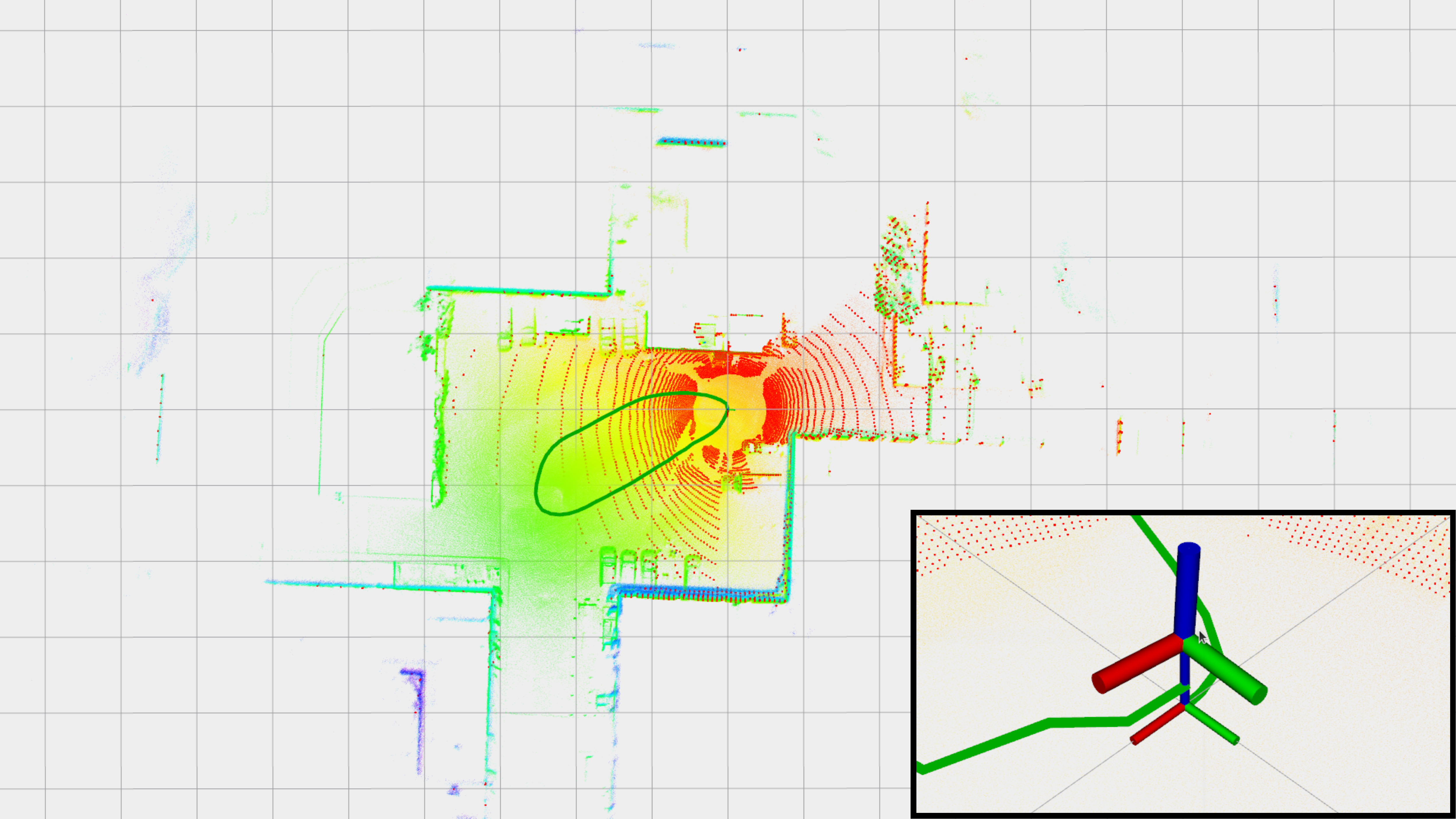}
    \end{subfigure}
    \hfill
    \begin{subfigure}[b]{0.245\linewidth}
        \centering
        \includegraphics[width=\linewidth]{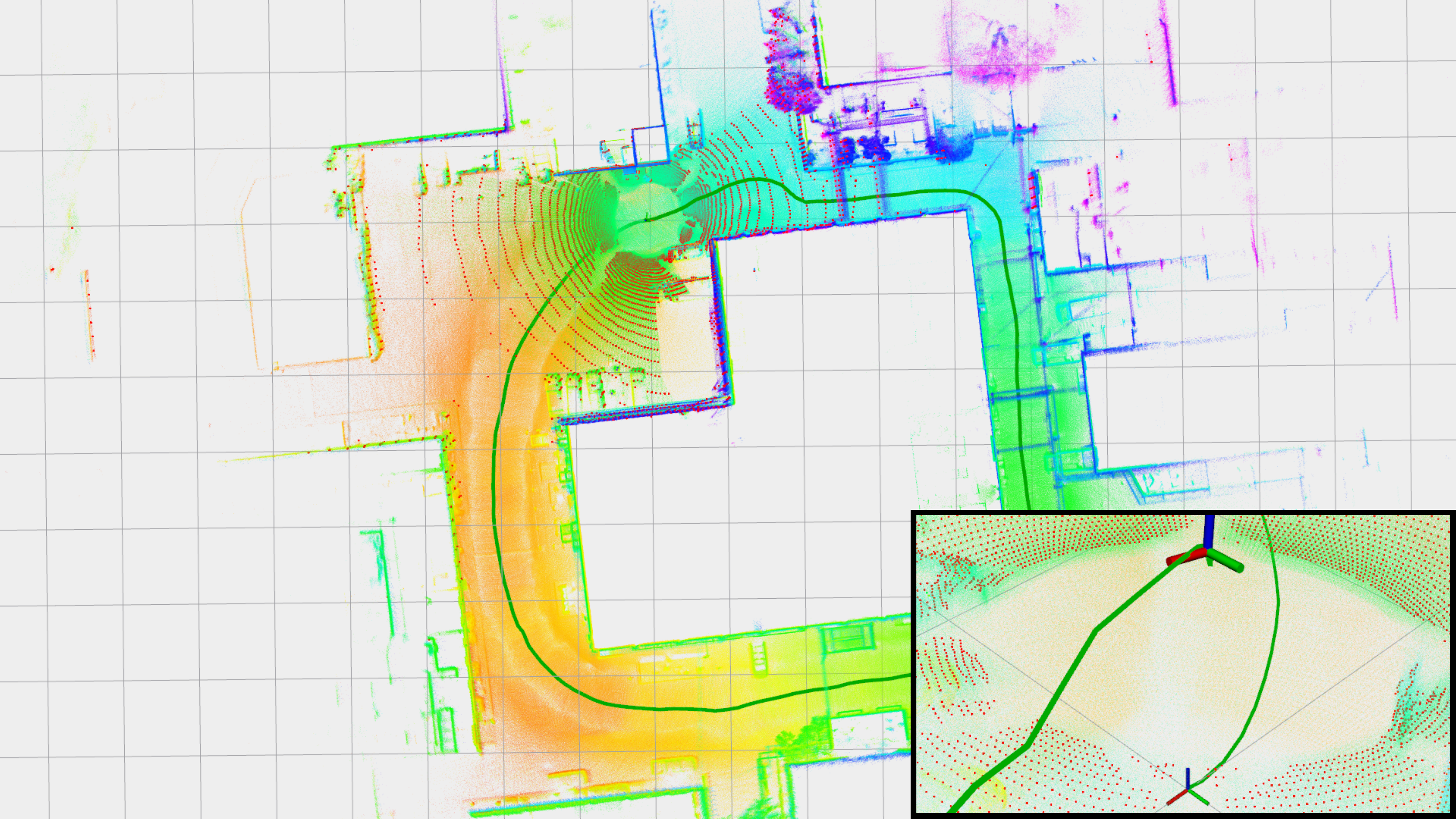}
    \end{subfigure}
    \hfill
    \begin{subfigure}[b]{0.245\linewidth}
        \centering
        \includegraphics[width=\linewidth]{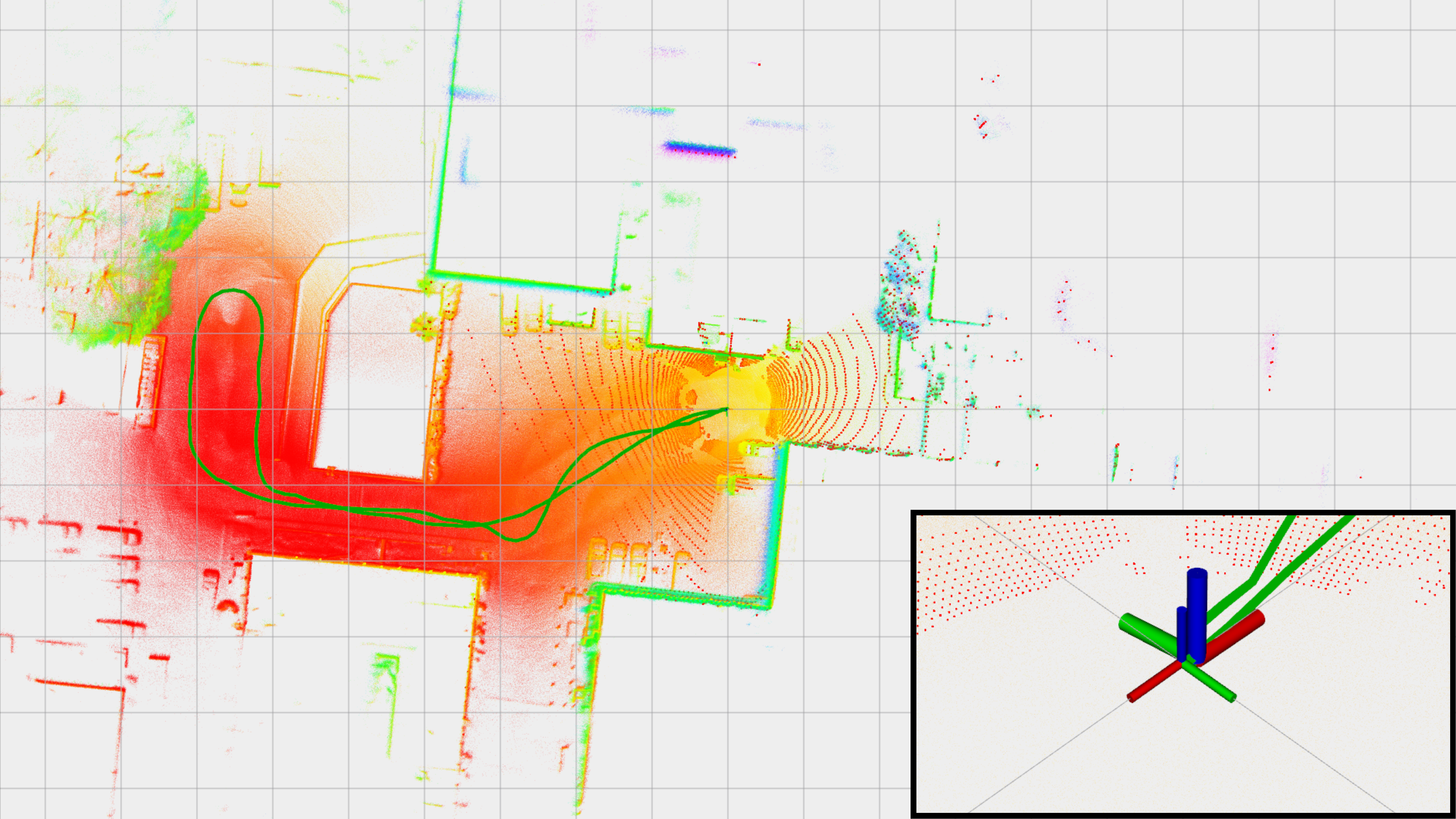}
    \end{subfigure}
    \hfill
    \begin{subfigure}[b]{0.245\linewidth}
        \centering
        \includegraphics[width=\linewidth]{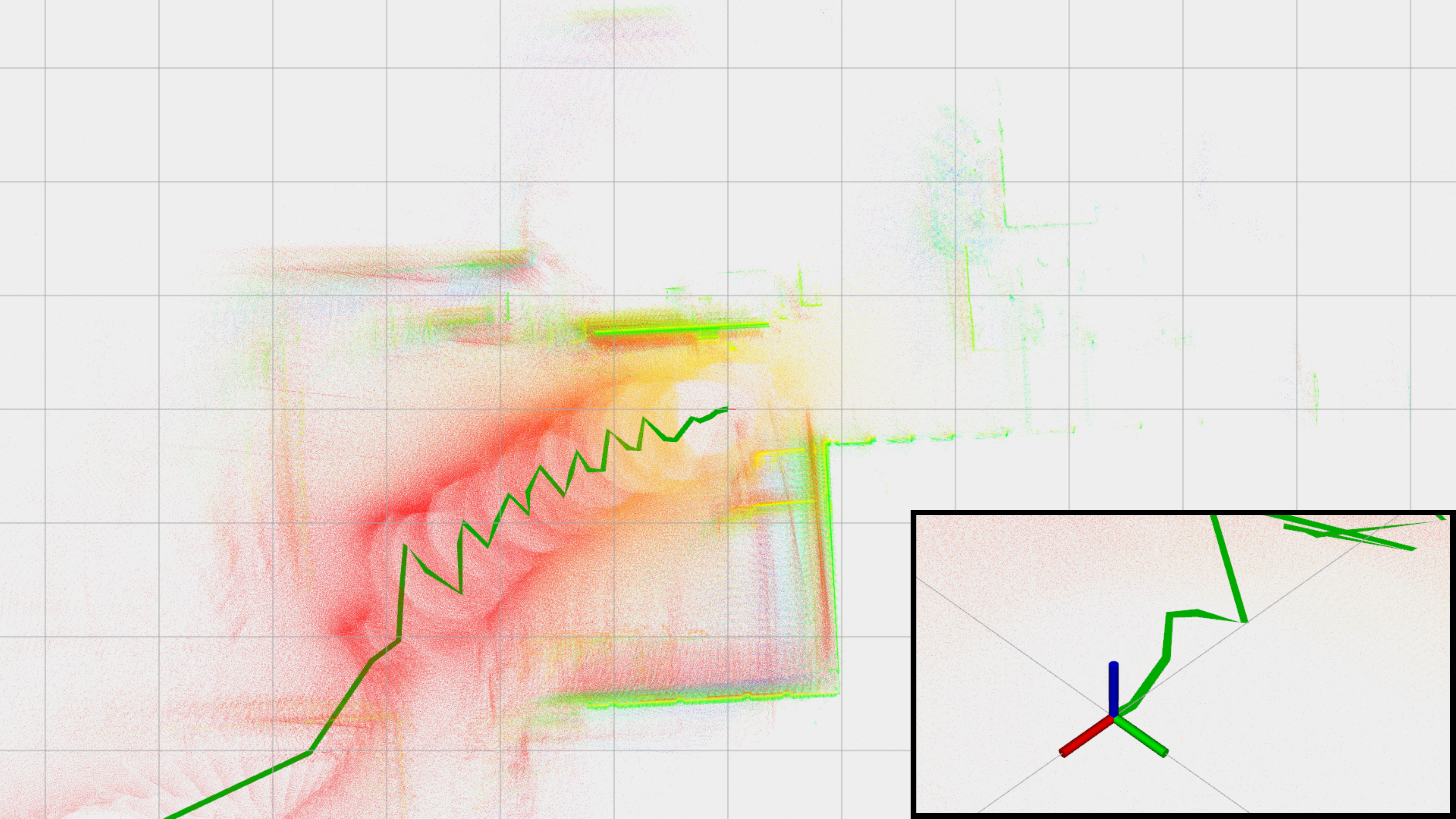}
    \end{subfigure}
    \hfill
    \begin{subfigure}[b]{0.245\linewidth}
        \centering
        \includegraphics[width=\linewidth]{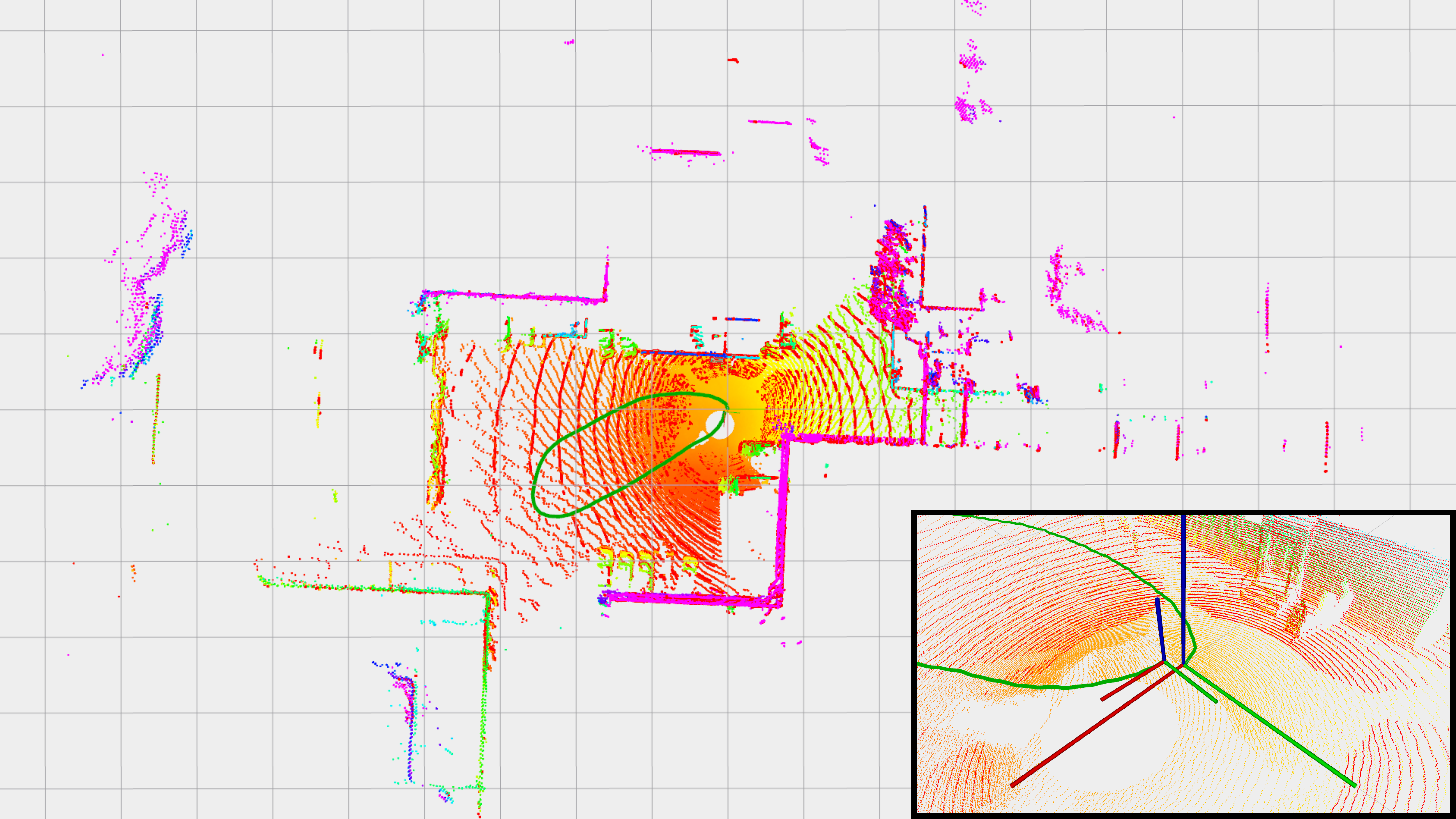}
        \caption{Seq 1: small loop}
    \end{subfigure}
    \hfill 
    \begin{subfigure}[b]{0.245\linewidth}
        \centering
        \includegraphics[width=\linewidth]{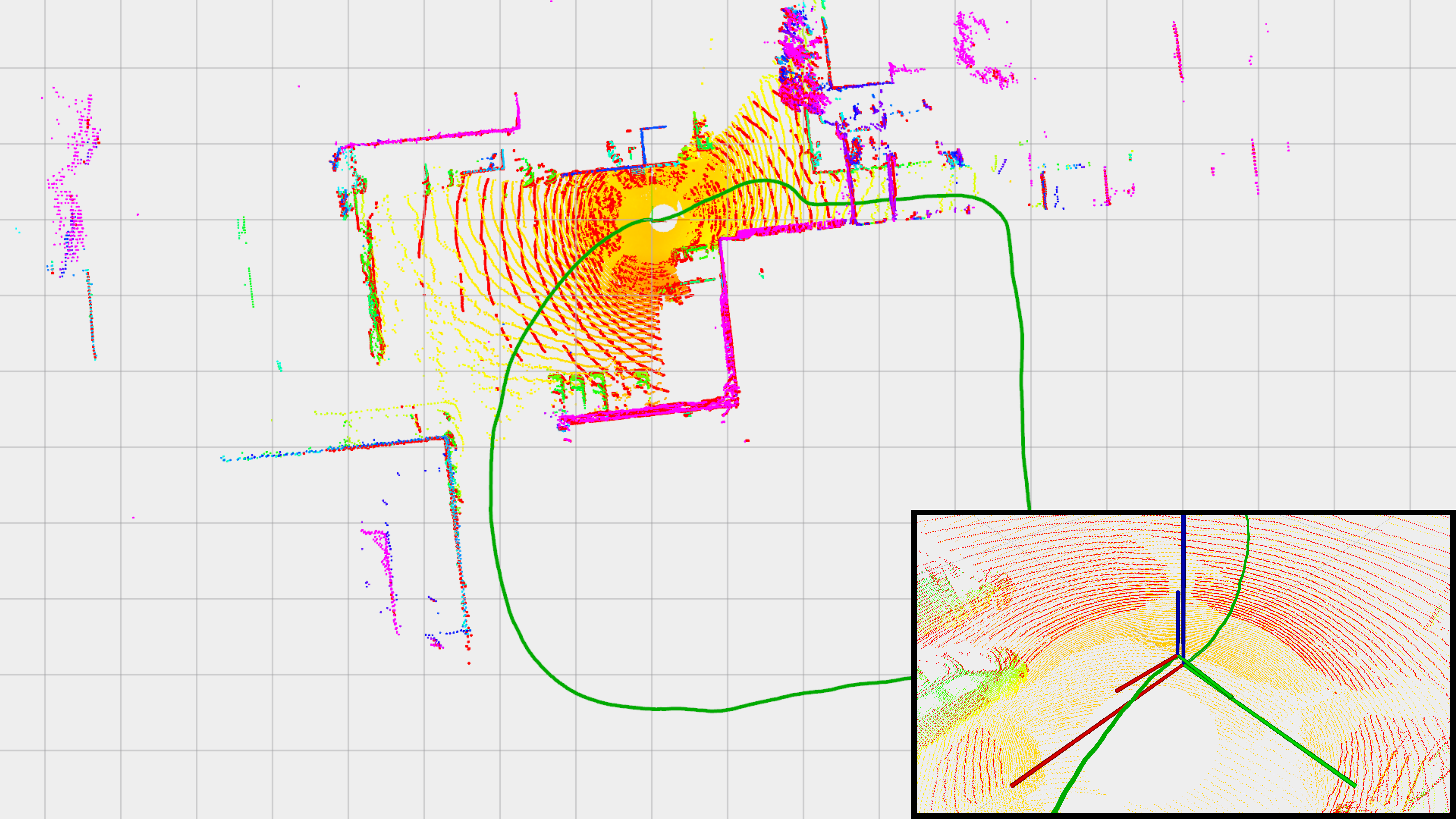}
        \caption{Seq 2: large loop}
    \end{subfigure}
    \hfill
    \begin{subfigure}[b]{0.245\linewidth}
        \centering
        \includegraphics[width=\linewidth]{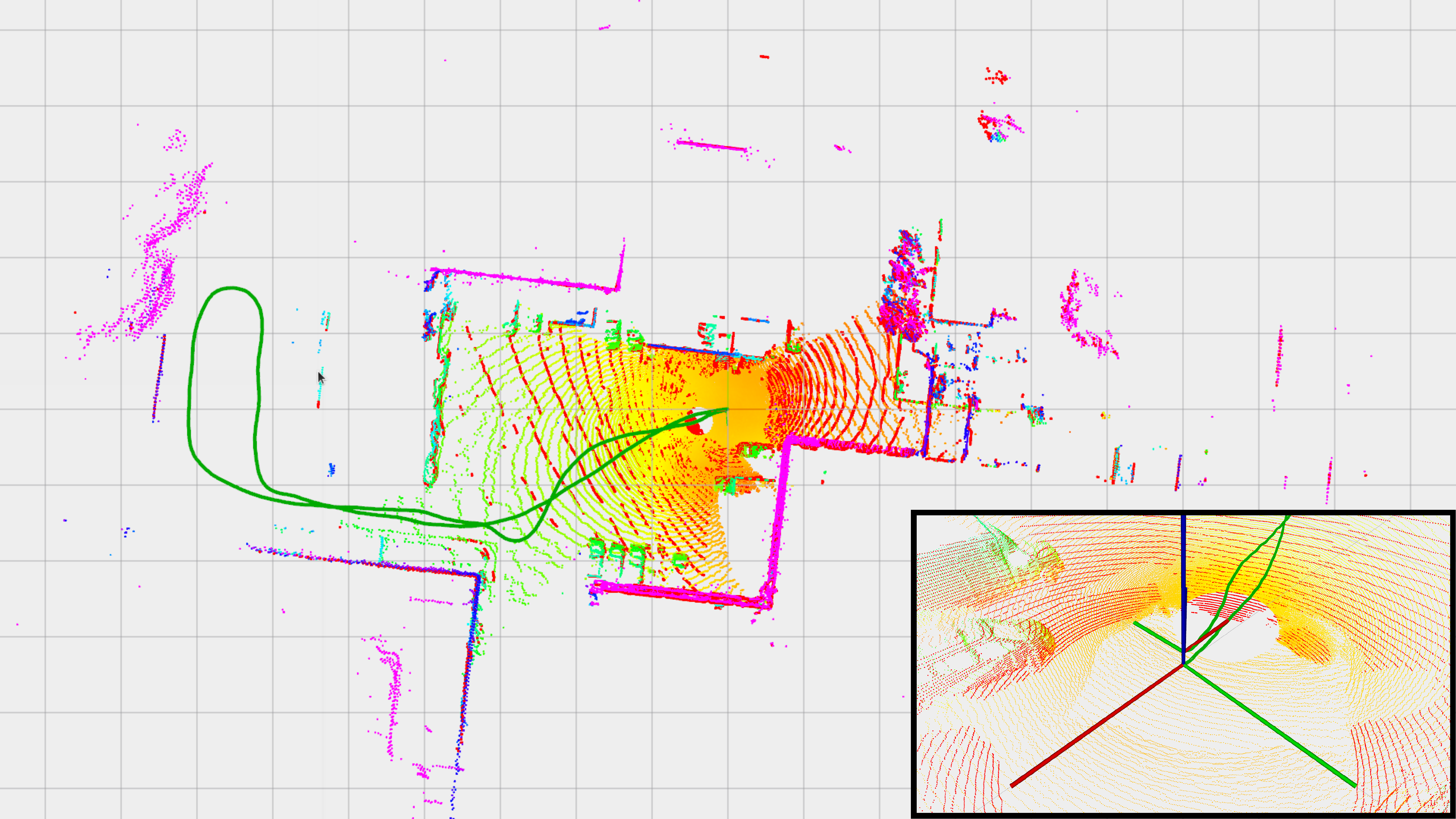}
        \caption{Seq 3: open space}
    \end{subfigure}
    \hfill
    \begin{subfigure}[b]{0.245\linewidth}
        \centering
        \includegraphics[width=\linewidth]{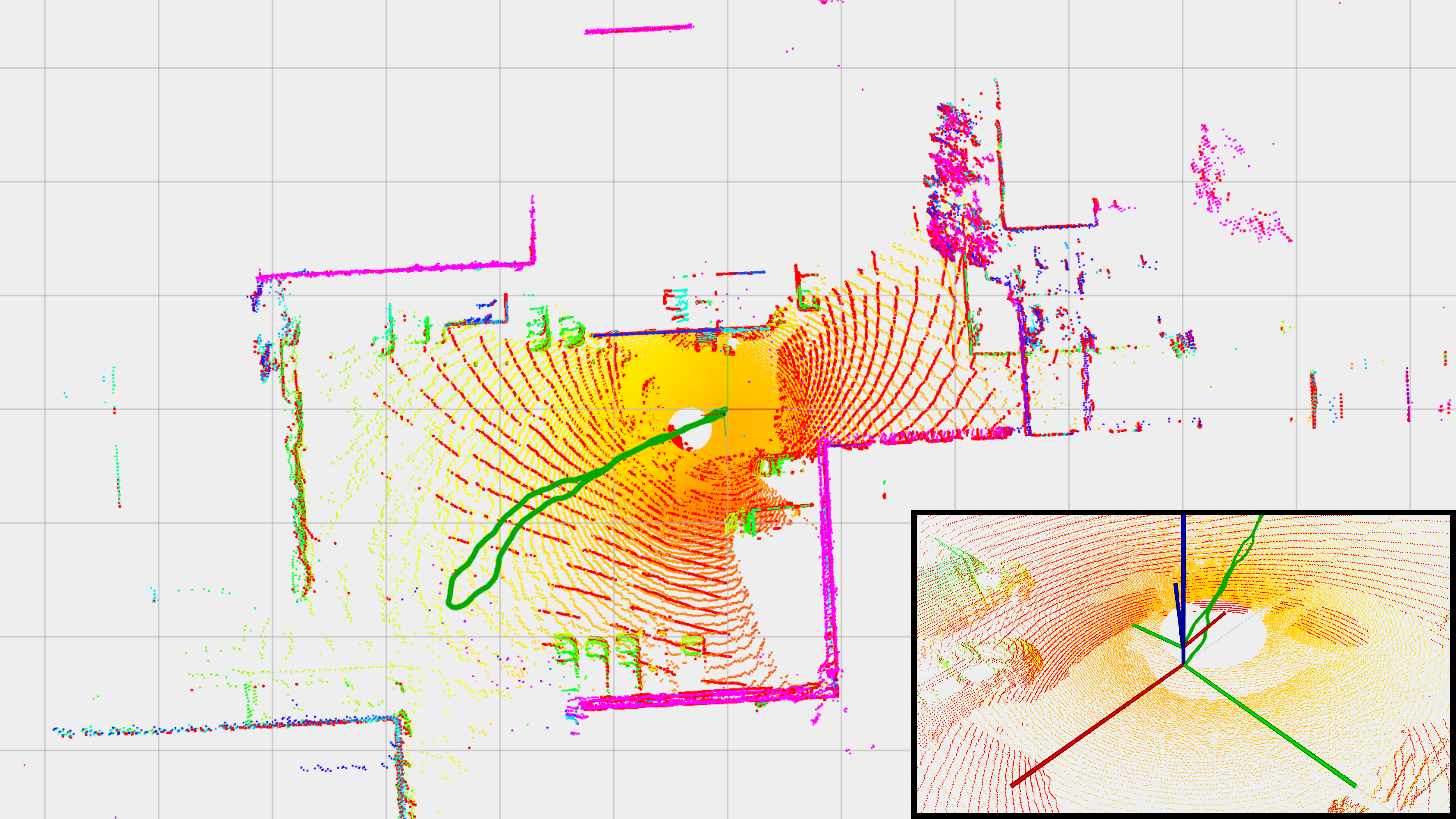}
        \caption{Seq 4: aggressive motion}
    \end{subfigure}
\caption{Qualitative results of \flio~(top) and \llol~(bottom) on our own datasets.}
\label{fig:perch}
\vspace{-8px}
\end{figure*}
\begin{table}[t]
\centering
\caption{Trajectory drift of \flio~and \llol~on our own dataset.}
\label{tab:perch}
\begin{tabular}{ccccccc}
\hline
  & \multicolumn{3}{c}{\flio} & \multicolumn{3}{c}{\llol} \\
Seq. & Distance & Error   & Drift    & Distance   & Error & Drift  \\
\hline \hline
1 & 69.53    & 0.91    & 1.31\%   & 64.88   & 0.38  & \textbf{0.58\%} \\
2 & 257.52   & 6.53    & 2.54\%   & 244.00  & 0.25  & \textbf{0.10\% }\\
3 & 208.20   & 0.21    & \textbf{0.10\%}   & 201.00  & 0.52  & 0.26\% \\
4 & 869.87   & 358.1   & 41.2\%   & 63.07   & 0.43  & \textbf{0.68\%} \\
\hline
\end{tabular}
\vspace{-8px}
\end{table}

In addition to the Newer College dataset, 
we also collect our own for evaluation.
We acquire the dataset with a handheld Ouster OS1 Gen2 sensor running at \SI{10}{Hz} with 1024 firings per sweep.
Following~\cite{Engel2016APC}, we come back to the starting point after a loop, 
and the difference between the first and the last pose can be used to evaluate odometry drift without ground truth.
We test in four distinct scenarios: 
a small loop, a large loop, an open space trajectory and an aggressive motion trajectory.
Quantitative results are shown in Table~\ref{tab:perch}.
We report the total distance estimated by each system and their final positional error.
The drift is calculated as the quotient of the two.
Both systems work well on the first 3 sequences with normal walking motion, 
but \flio~failed to converge on the last one with aggressive rotation.
\llol~is able to achieve $<$1\% drift on all sequences.
Qualitative results are shown in Fig.~\ref{fig:perch}.
\subsection{Runtime and Latency}

We benchmark both \llol~and \flio~on a desktop CPU (Intel i7-6700K@4.2GHz) 
as well as an embedded one (ARMv8.2@2.2GHz, NVIDIA Xavier) under single and multi-threaded settings.
In addition, we conducted experiments on both indoor and outdoor datasets.
The outdoor environment is roughly twice the size of the indoor one.
This is to demonstrate the effect of workspace sizes on each system.
For \flio, we report the total time in the same way as the original paper.
For \llol, we record time from the (partial) sweep arrival to the end of pano update.
The pano rendering only runs infrequently in the background and is not included in the timing results.

\begin{figure}
\centering
\includegraphics[width=\linewidth]{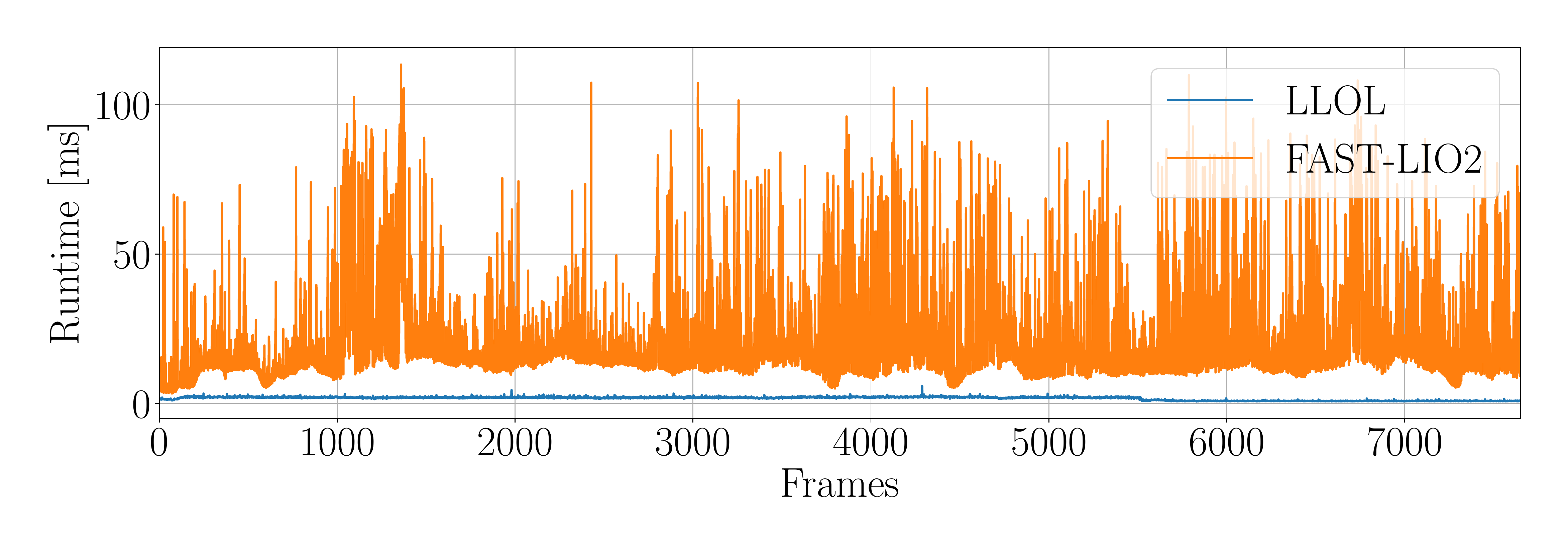}
\caption{Per-frame runtime of \flio~and \llol~on NC01, 
measured on Intel CPU with 8 threads.
\llol~has an average runtime of 1.70$\pm$\SI{0.55}{ms}.
\flio~has an average runtime of 21.73$\pm$\SI{16.50}{ms}.}
\label{fig:nc01-runtime}
\vspace{-8px}
\end{figure}

Runtime on our own datasets are shown in Table~\ref{tab:runtime}. 
We see that \llol~is in general much faster than \flio, 
with sub-millisecond processing time in some cases.
\flio~exhibits large variance in runtime between indoor and outdoor environments,
while \llol~runs in near constant time regardless of workspace size.
We also visualize the per-frame runtime on NC01 in Fig.~\ref{fig:nc01-runtime}.
While \llol~maintains a fairly stable performance over the entire sequence, 
\flio~displays inconsistent results, with intermittent bursts up to 100ms.
The latency of each method can be computed by adding the sensor integration time with the runtime.
For example, running \llol~with a quarter sweep will result in an average latency of \SI{28.37}{ms} on Xavier with a single thread, 
as opposed to \SI{189.81}{ms} for \flio.

Finally, ours system running on embedded CPUs compares favourably to UPSLAM~\cite{Cowley2021UPSLAMUO} (\SI{5}{ms} on Xavier) which requires a GPU.
LITAMIN2~\cite{Yokozuka2021LiTAMIN2UL} (\SI{5}{ms} on Intel CPU) can also achieve similar speed as ours,
but the lack of open source implementations precludes a fair comparison.

\begin{table}[t]
\centering
\caption{Runtime comparison of \flio~vs \llol~on different types of processors
under single and multi-threaded settings for both indoor (top) and outdoor (bottom) environments.
Results are averaged within the entire sequence.}
\label{tab:runtime}
\begin{tabular}{cccccc}
\hline
\multicolumn{2}{c}{Runtime [ms]} & \multicolumn{2}{c}{Xavier (ARMv8.2)} & \multicolumn{2}{c}{Intel (i7-6700K)} \\
method & sweep (ms)  & single & multi (8) & single & multi (8) \\
\hline \hline
\flio  & full (100) & 44.59 & 27.13 & 22.22 & 21.09\\
\llol  & full (100) & 12.56 & 4.46 & 5.24 & 1.93 \\
\llol  & $1/2$ (50) & 6.87 & 3.02 & 3.02 & 1.29 \\
\llol  & $1/4$ (25) & \textbf{4.32} & \textbf{2.30} & \textbf{1.87} & \textbf{0.94} \\
\hline
\flio  & full (100) & 89.81 & 34.40 & 42.67 & 28.63 \\
\llol  & full (100) & 10.15 & 3.77 & 3.89 & 1.56 \\
\llol  & $1/2$ (50) & 5.68 & 2.55 & 2.33 & 1.07 \\
\llol  & $1/4$ (25) & \textbf{3.37} & \textbf{1.89} & \textbf{1.46} & \textbf{0.77}  \\
\hline
\end{tabular}
\vspace{-8px}
\end{table}

\section{Conclusions}
In this paper, we present a low-latency odometry system for spinning lidars.
It has comparable accuracy and drift to a state-of-the-art lidar odometry system~\cite{Xu2021FASTLIO2FD},
but runs an order of magnitude faster.
The decision to handle partial sweeps significantly boosts the odometry update frequency of our system compared to approaches operating at the nominal rotational rate of the lidar sensor.
A key enabler of our fast implementation is the use of depth panoramas as our local map representation, 
this affords us constant time search and update complexity across different environment scales.
Our odometry system offers a high data throughput combined with a low memory footprint,
these features make it suitable for deployment on resource constrained platforms.
We open source our implementation as a contribution to the community.


\bibliographystyle{IEEEtran}
\bibliography{IEEEabrv,reference}

\end{document}